\documentclass[sigconf]{acmart}
\usepackage{amsmath,amsfonts}
\usepackage{adjustbox}

\usepackage{balance}
\usepackage{appendix}
\usepackage{algorithm}
\usepackage[noend]{algpseudocode}
\usepackage{graphicx}
\usepackage{textcomp}
\usepackage{xcolor}
\usepackage{caption}
\usepackage{subcaption}
\usepackage{tabularx}
\usepackage{multirow}
\usepackage{makecell}
\usepackage{diagbox}
\usepackage{booktabs}
\usepackage{color, colortbl}
\usepackage{courier}
\usepackage{pifont}
\definecolor{Gray}{gray}{0.9}
\algrenewcommand\algorithmicrequire{\textbf{Input:}}
\algrenewcommand\algorithmicensure{\textbf{Output:}}

\AtBeginDocument{%
 \providecommand\BibTeX{{%
  Bib\TeX}}}

\author{Yoonhyuk Choi}
\orcid{0000-0003-4359-5596}
\affiliation{
  \institution{
  Samsung}
  \city{Seoul}
  \country{Republic of Korea}
}
\email{chldbsgur123@gmail.com}

\author{Jiho Choi}
\orcid{0000-0002-7140-7962}
\affiliation{
  \institution{
  KAIST}
  \city{Seoul}
  \country{Republic of Korea}
}
\email{jihochoi @ kaist.ac.kr}

\author{Taewook Ko}
\orcid{0000-0001-7248-4751}
\affiliation{
  \institution{
  Samsung}
  \city{Seoul}
  \country{Republic of Korea}
}
\email{taewook.ko @ snu.ac.kr}

\author{Chong-Kwon Kim}
\orcid{0000-0002-9492-6546}
\affiliation{
  \institution{Korea Institute of Energy Technology}
  \city{Naju}
  \country{Republic of Korea}
}
\email{ckim @ kentech.ac.kr}
\renewcommand{\shortauthors}{Yoonhyuk Choi, Jiho Choi, Taewook Ko, and Chong-Kwon Kim}

\newtheorem{manualtheoreminner}{\textbf{Theorem}}
\newenvironment{manualtheorem}[1]{%
  \renewcommand\themanualtheoreminner{\textbf{#1}}%
  \manualtheoreminner
}{\endmanualtheoreminner}

\begin{document}

\title{Mitigating Overfitting in Graph Neural Networks via Feature and Hyperplane Perturbation}

\begin{abstract}
Message-passing neural networks have shown satisfactory performance and are widely employed in various graph mining applications. However, like many other machine learning techniques, these methods suffer from the overfitting problem when labeled data are scarce. Our observations suggest that sparse initial vectors along with scarce labeled data further exacerbate this issue by failing to fully represent the range of learnable parameters. 
%This sparsity can hinder the optimization of specific dimensions in the initial projection matrix, as the training samples may not adequately span these parameters. 
To overcome this challenge, we propose a novel perturbation technique that introduces variability to the initial features and the projection hyperplane. 
%Notably, even without employing grid search, we demonstrate that shifting with a small estimated value mitigates this problem more effectively than other perturbation methods. 
Experimental results on real-world datasets reveal that our technique significantly enhances node classification accuracy in semi-supervised scenarios. In addition, our extensive experiments reveal the effect and robustness of hyperplane shifting.
\end{abstract}

\begin{CCSXML}
<ccs2012>
   <concept>
       <concept_id>10003752.10010070.10010071.10010289</concept_id>
       <concept_desc>Theory of computation~Semi-supervised learning</concept_desc>
       <concept_significance>500</concept_significance>
       </concept>
 </ccs2012>
\end{CCSXML}

\ccsdesc[500]{Theory of computation~Semi-supervised learning}

\keywords{Graph neural networks, semi-supervised learning, feature sparsity}

\maketitle

\begin{figure}[t]
    \centering
    \includegraphics[width=.5\textwidth]{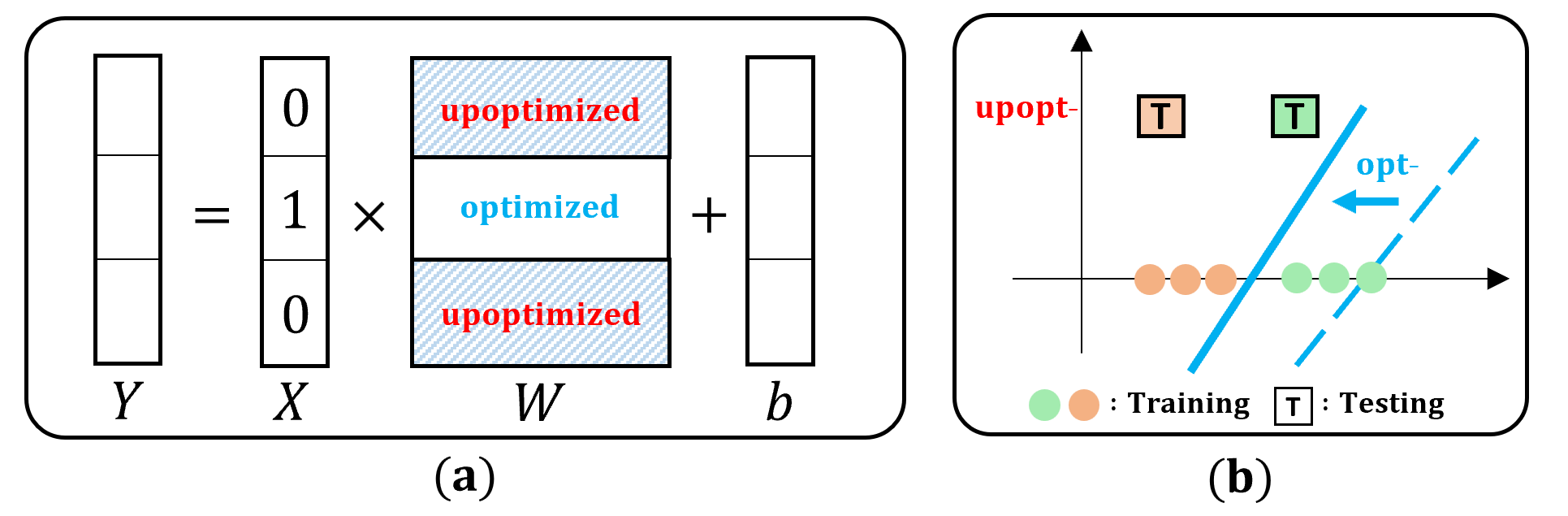}
    \caption{The colored area in $W$ indicates that the gradient is not updated during back-propagation, as it depends on whether the input element $X$ is zero. Consequently, the parameter is optimized only for the x-axis during training}
    \label{example}
\end{figure}

%\begin{figure}[t]
%    \centering
%    \includegraphics[width=.5\textwidth]{assets/example.png}
%    \caption{In the left figure (a), the colored area in $W$ indicates that the gradient is not updated during back-propagation, as it depends on whether the input element $X$ is zero. Consequently, in the right figure (b), the parameter is optimized (opt-) only for the x-axis during training}
%    \label{example}
%\end{figure}

\section{Introduction}
As graphical data are accumulated and accessible publicly, Graph Neural Networks (GNNs) have attracted huge research attentions and many clever algorithms have been proposed. They are successfully applied in diverse downstream tasks including molecular prediction and social network analysis. Recent studies \cite{defferrard2016convolutional,kipf2016semi,hamilton2017inductive,velickovic2017graph} underline their effectiveness in node and graph classification tasks, which largely results from the seamless integration of node features and network structures. As a core component of GNNs, message-passing enables information exchange between neighboring nodes and updates the features through iterative processes \cite{gilmer2017neural}.

Message passing is particularly effective on homophilic graphs where connected nodes typically share similar labels \cite{mcpherson2001birds}. However, performance of GNNs drops significantly on heterophilic graphs where neighboring nodes often have differing labels. Several researchers have proposed novel schemes to rectify the problem. One class of schemes focuses on the aggregation of features passed from neighbor nodes. They adjust edge weights before aggregation \cite{velickovic2017graph, yang2019masked, bo2021beyond, kim2022find} or even eliminate edges that exhibit disassortative traits \cite{ying2019gnnexplainer, luo2021learning}. Other approaches utilize distant but similar nodes \cite{pei2020geom, jin2021node, li2022finding} or employ node-specific propagation techniques with adaptable parameters \cite{xiao2021learning}. This emphasizes the importance of selecting appropriate aggregation strategies but raises a fundamental question: \textit{are there other factors at play beyond aggregation schemes?}

To address this issue, our method departs from common practices by focusing on optimal training of learnable weight matrices (hyperplanes) rather than relying solely on aggregation. This approach is motivated by the vulnerability of sparse initial features, such as bag-of-words representations, especially in semi-supervised settings with limited training samples. Sparse initial vectors often lead to overfitting in certain dimensions of the initial layer’s parameters, ultimately reducing prediction accuracy. The problem arises when test nodes exhibit feature dimensions that are insufficiently represented during the training phase, resulting in incomplete learning.

Figure \ref{example} illustrates a representative case of this problem in detail. Specifically, Figure \ref{example}a displays the input data $X$, the trainable weight matrix $W$, and its associated bias $b$. Using this configuration, we define a loss function $\mathcal{L}$ to adjust the position of the weight matrix $W$ as $\bigtriangledown_{W}\mathcal{L}=-(Y-\widehat{Y})X$, where $\mathcal{L}={1 \over 2}(Y-\widehat{Y})^2$ and $\widehat{Y}=W^TX+b$. The above equation calculates the partial derivative ($\bigtriangledown$) relative to the learnable parameters. Given that many components in the input feature vector are zero, such as $X=(0,1,0)$, the gradient plane $W$ only updates the second dimension. In this setup, the bias adjusts to predict the appropriate label accordingly. Even though the parameter is optimized for accurate predictions, we find that the prevalence of zero components becomes increasingly influential in high-dimensional spaces. Let us assume that the initial vector is sparse so that only the $i$-th component is non-zero.
%As shown in Figure \ref{example}b, we examine a case where the training features (indicated in color) are mostly zeros, aside from those on the $x$-axis. 
Then, only the gradients along the $i$-axis of the plane undergo updates. This condition presents significant challenges for classifying test nodes, a difficulty that becomes even more pronounced in high-dimensional spaces with numerous zero-valued features. This is a phenomenon that is widely recognized as overfitting.

\begin{figure}[t]
    \centering
    \includegraphics[width=.48\textwidth]{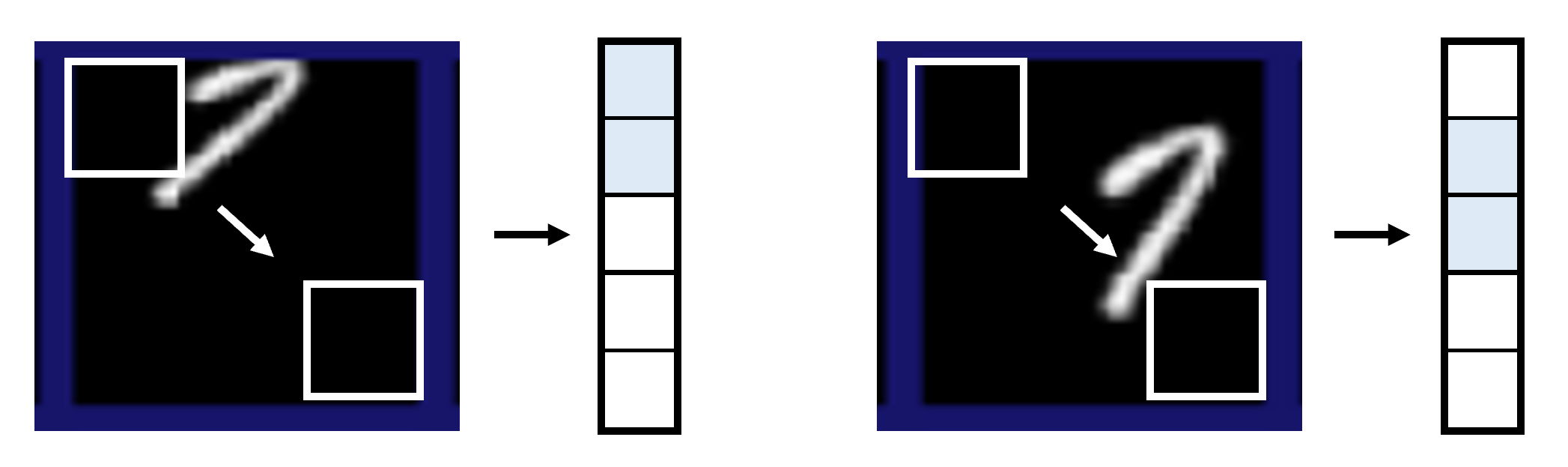}
    \caption{Translation invariance of convolutional networks. The white square represents the sliding convolution filter, while the blue squares indicate non-zero outputs}
    \label{mnist}
\end{figure}

For the proper processing of the initial features at the first-layer, eliminating or reducing zero components becomes imperative. Our prior investigation led us to consider data augmentation techniques like dimensional shifting, widely applied in the field of computer vision \cite{shijie2017research}. Nevertheless, we determined that such methods were ill-suited to GNNs employing bag-of-words features, as they might compromise the integrity of semantic information. Furthermore, as shown in Figure \ref{mnist}, multilayer perceptrons (MLP) within GNN lack the translation invariance characteristic of convolutional neural networks \cite{zhang2017rotation}.
A perturbation scheme which introduces noise to the inputs can be contemplated \cite{zhu2021graph}, however, it provokes its own set of complications including the necessity for meticulous parameter tuning and normalization \cite{cai2021graphnorm}.

We propose a solution that combines shifting of the initial features and parameters simultaneously. This approach can preserve the volume of gradients and reduce the variance of predictions, which is built upon previous techniques such as parameter shifting \cite{kim2019zero} and rotation of neural networks \cite{lin2020rotated}. The proposed scheme is also related to a dual-path network \cite{chen2017dual} that enables operations in both the original and shifted spaces. Consequently, our method addresses the zero gradient problem caused by input data and promotes accurate semantic learning in entire dimensions \cite{liu2022confidence}. 

We claim that the proposed perturbation mechanism is orthogonal to GNNs, which can be applied to various representative algorithms such as MLP, GCN \cite{kipf2016semi}, GAT \cite{velickovic2017graph}, and FAGCN \cite{bo2021beyond}. We compared the performance of shifting variants (Shift-MLP, Shift-GCN, Shift-GAT, and Shift-FAGCN) with the original methods and observed performance gains of 14.0\%, 16.8\%, 13.1\%, and 10.4\%, respectively. Lastly, we confirm that our approach surpasses all existing state-of-the-art methods on nine datasets with bag-of-word features. We performed extensive sensitivity analysis and ablation study. Our experiments confirm the effectiveness and reliability of the proposed shifting algorithm. 
%Additional experimental results can be found in our full paper \cite{choi2022perturb}. 
Summarizing the above perspectives, our contributions can be described as follows:

\begin{itemize}
\item We first identify that GNN training is highly sensitive to the sparsity of initial feature vectors, particularly in bag-of-words representations \cite{choi2022perturb}. 
%We point out some misunderstandings (flip-based learning) in \cite{choi2022perturb} and demonstrate how shift-based adjustments can significantly improve model performance. 
To ensure robust feature representation, we introduce a novel co-training approach that concurrently shifts both initial features and the hyperplane, focusing on gradient flow and back-propagation rather than solely on aggregation. %This approach provides targeted, component-wise guidance for the initial projection matrix. 
\item We provide experimental results that indicate the shifting mechanism functions independently of message-passing schemes, enabling integration with state-of-the-art GNN algorithms. Comprehensive evaluations of nine benchmark datasets demonstrate that our shift-based variants significantly outperform all existing methods.
\item We analyze the proposed algorithm via extensive sensitivity analysis and ablation study. Our results confirm the robustness of the proposed algorithm. We also scrutinize the effects of first hyperplane shifting in a greater detail. 
\end{itemize}

\section{Related Work}
Previous message-passing algorithms generally fall into two main categories: spectral-based and spatial-based approaches \cite{pasa2021multiresolution, song2022graph, waikhom2022recurrent}. Spectral-based GNNs provide a theoretical framework for graph convolutions within the spectral domain by leveraging the Laplacian matrix \cite{defferrard2016convolutional, dong2021graph, bianchi2021graph}. In contrast, spatial-based GNNs perform aggregation from local node neighborhoods in Euclidean space, spurring the development of diverse aggregation techniques to address noisy or heterophilic edges \cite{velickovic2017graph, pei2020geom, luan2022revisiting, wang2022powerful, peng2022reverse, li2022finding, liu2023comprehensive, wu2023beyond}. While these methods have advanced the edge of research in GNN, they primarily concentrate on refining message-passing mechanisms. Consequently, the problem caused by the sparsity in initial feature representations remains largely unexplored in current literature.

In addition to message-passing, numerous recent approaches have been proposed to enhance neural network generalization \cite{chen2017stochastic, feng2017sparse, wang2021confident, yang2022ncgnn, fan2023generalizing, wang2023fl}. Some of these methods focus on normalizing deep neural networks \cite{huang2020normalization}, while others employ regularization across adjacent nodes \cite{yang2021rethinking}, integrate label propagation for additional information \cite{wang2020unifying}, or address data sparsity issues \cite{zhou2021sparsity, ye2021sparse, cheng2023graph, jiang2023graph}. More recently, orthogonal constraints have been introduced to counteract the gradient vanishing problem in the early layers of GNNs \cite{kim2022revisiting, guo2022orthogonal, esguerra2023sparsity}. Additionally, RawlsGCN \cite{kang2022rawlsgcn} highlights the issue of biased gradient updates, favoring nodes with higher degrees. These methods have demonstrated significant improvements in semi-supervised scenarios; however, they do not fully address challenges stemming from the intrinsic characteristics of initial features. In this paper, we tackle this issue by simultaneously shifting both features and parameters.

\section{PRELIMINARIES}
This section starts with an empirical analysis to give an overview of the feature distribution across six benchmark graph datasets. Following that, we provide the commonly used notations in Graph Neural Networks (GNNs). Finally, we explain the mechanism of GNNs in terms of feature projection and message-passing.

\begin{table}[t]
\caption{The $z$-value of datasets: Homophilic (Cora, Citeseer, PubMed) and Heterophilic (Cornell, Texas, Wisconsin). 
More details of these datasets are provided in Table \ref{dataset}}
\begin{adjustbox}{width=.48\textwidth}
\begin{tabular}{@{}ll|c|c|c|c|c|c}
\Xhline{2\arrayrulewidth}
        & Datasets 
        %& \multirow{2}{*}{Datasets}                       %& \multicolumn{3}{c}{Range of the node sets (S)} \\
%\cline{3-5}
        & Cora  & Cite. & Pubm. & Corn. & Texas & Wisc.\\ 
\Xhline{2\arrayrulewidth}
                        & $z$-value              & 0.59 & 0.42 & 0.96 & 0.63 & 0.41 & 0.52 \\
\Xhline{2\arrayrulewidth}
\label{status}
\end{tabular}
\end{adjustbox}
\end{table}

\subsection{Statistics on Benchmark Graphs}
In Table \ref{status}, we present the \(z\) value, which is the ratio of non-zero components in the initial feature vector. 
%variation of the variable \(z\) while altering the range of the node set \(S\) from the ego to its 2-hop neighboring nodes. Given the initial feature set \(S \in \mathcal{R}^F\), we define a vector $m$ as the addition of the node feature vectors (\(X_v\)) as below:
%\begin{equation}
%    m = \sum_{v \in S}{X_v}
%\end{equation}
%The $z$ value, the ratio of non-zero elements in $m$, is determined as 
%\begin{equation}
%\label{nz_ratio}
%z = 1-\frac{\sum^{F}_{i=1}\delta(m_i, 0)}{F}
%\end{equation}
%where \(\delta(m_i, 0) = 1\) if the \(i^{\text{th}}\) element in \(m\) is 0. 
%This allows \(z\) to be obtained by dividing the numerator by the dimension of the feature vector \(F\). 
%As shown in Table \ref{status}, the value of \(z\) tends to increase as the range expands, primarily due to the availability of more features. 
It is important to note that the scale of $z$ varies significantly across different graphs, depending on the specific characteristics of the input data (refer to the provided dataset links in $\S$ \ref{data_desc}). In summary, a lower value of $z$ corresponds to a notable performance improvement, with a theoretical analysis focusing on gradient updates and variance reduction provided in $\S$ \ref{theo_anal}.
We can extend the definition of the $z$ value from an ego node to neighbor node sets, but omit the extended definition because these values are proportional to the value of ego nodes.

\subsection{Definition of Notation}
Let $\mathcal{G} = (\mathcal{V}, \mathcal{E})$ be an undirected graph consisting of $n = \vert\mathcal{V}\vert$ nodes and $m = \vert\mathcal{E}\vert$ edges. The graph structure is represented as adjacency matrix $A \in \{0,1\}^{n \times n}$. Each node is characterized by a feature matrix $X \in \mathbb{R}^{n \times F}$, where $F$ represents the dimensionality of the initial features. Node labels are denoted by $Y \in \mathbb{R}^{n \times C}$, with $C$ indicating the number of classes. This study aims to tackle a node classification task in a semi-supervised context. Specifically, we aim to enhance the utilization of the given feature matrix $X$ to predict the labels of the unlabeled nodes $\mathcal{V}_U ,(= \mathcal{V} - \mathcal{V}_L$), when only a subset of labeled nodes $\mathcal{V}_L \subset \mathcal{V}$ is available for training.

\subsection{Graph Neural Network (GNN)} \label{basic_gnn}
Spatial GNNs have shown significant performance enhancements in semi-supervised settings, while also reducing the computational burden of Laplacian decomposition. The basic framework of these models is described as follows:
\begin{equation}
\label{gnn}
\begin{gathered}
H^{(l+1)}=\sigma(AH^{(l)} W^{(l)})
\end{gathered}
\end{equation}
An adjacent matrix $A$ is used for message-passing. $H^{(1)}=X$ represents the initial node features, and $\bar{H}^{(l)}$ denotes the hidden representation at the \textit{l-th} layer. The transformation of $H^{(l)}$ is carried out using an activation function $\sigma$ (e.g., ReLU). GNNs generate the final prediction, $\bar{H}^{(L)}$, by applying a softmax function. In this framework, $W^{(l)}$ are the trainable weight matrices shared across all nodes and updated using the negative log-likelihood loss ($\mathcal{L}_{nll}$) between the predicted values $\widehat{Y}=\sigma(H^{(L)})$ and the true labels $Y$, as shown below:
\begin{equation}
\label{loss_gnn}
\mathcal{L}_{GNN}=\mathcal{L}_{nll}(Y, \widehat{Y})
\end{equation}
GNNs aim to improve aggregation mechanisms to create effective message-passing schemes \cite{velickovic2017graph, klicpera2018predict, chen2020simple, bo2021beyond}. %For example, GCN \cite{kipf2016semi} employs a normalized Laplacian matrix, GAT \cite{velickovic2017graph} constructs an aggregation matrix by calculating attention scores between nodes, and APPNP \cite{klicpera2018predict} incorporates personalized PageRank to optimize the propagation process. 
Nevertheless, updating the initial hyperplane based on input features has yet to be explored, which we will introduce below.

\subsection{Limitation of Sparse Features}
While effective aggregation schemes are undeniably essential for achieving optimal performance, as we will explain further, these schemes alone cannot address the problems arising from the sparse nature of the initial features. 
%To elaborate, we analyze the updates of the weight matrices $W^{(l)}$ as follows:
%\begin{equation}
%\begin{gathered}
%\label{gnn_grad}
%\bigtriangledown_{W^{(l)}}J=(AH^{(l)})^T\bigtriangledown_{\bar{H}^{(l+1)}}J,
%\end{gathered}
%\end{equation}
%where $J=\mathcal{L}_{GNN}$.
Intuitively, a lower value of $A$ can impede the gradient flow between dissimilar nodes.
Nevertheless, a problem emerges when updating the parameters of initial layer $W^{(1)}$ as below: 
\begin{equation}
\label{par_w1}
\begin{gathered}
\bigtriangledown_{W^{(1)}}J=(AX)^T\bigtriangledown_{\bar{H}^{(2)}}J,
%=(AX)^T\bigtriangledown_{\bar{H}^{(2)}}J
\end{gathered}
\end{equation}
where $J=\mathcal{L}_{GNN}$ and $X=H^{(1)}$ is the initial feature. The gradient of $W^{(1)}$ can be simply derived by differentiating $J$ concerning $\bar{H}^{(2)}$, where the value of $AX$ influences the magnitude of this gradient. In this context, certain dimensions with zero input values result in zero gradients.
%($\forall_i \in F=0 \Rightarrow \bigtriangledown_{\bar{H}_i^{(2)}}J=0$). 
Consequently, the update of $W^{(1)}$ is governed by the sparseness of the input features. In addition, the limited availability of training samples in semi-supervised scenarios exacerbates the problem of improper optimization. our approach focuses on removing zero components in $X$ to enable $W^{(1)}$ to learn the significance of each dimension accurately.

A straightforward approach to removing zero components from the initial features $X$ involves shifting, which can be simply achieved by adding a small-valued vector ($\varepsilon$) to $X$:
\begin{equation} \label{shifting} 
    X_s = X + \varepsilon 
\end{equation}
We evaluate the node classification accuracy of a two-layer GCN \cite{kipf2016semi} and compare it with GCN+S that uses shifted features ($X_s$) as inputs. When the shift value was small (0.1 to 0.2), a performance improvement was observed (GCN: 79.0\% $\rightarrow$ GCN+shift: 81.5\%). However, as $\varepsilon$ increased, the performance declined sharply. For example, with $\varepsilon = 1.0$, the accuracy of the shifted GCN decreased to nearly 30. 0\%. This reduction in performance is attributed to the non-shift-invariant nature of the projection matrix, which can potentially affect robustness \cite{singla2021shift}. Additionally, shifting alters the input magnitude, an important factor in normalizing neural networks. To address this, we propose adjusting the features and the first hyperplane simultaneously. This approach can preserve the invariance property and has a potential to improve prediction variance.

\begin{figure}[t]
\centering
    \includegraphics[width=.49\textwidth]{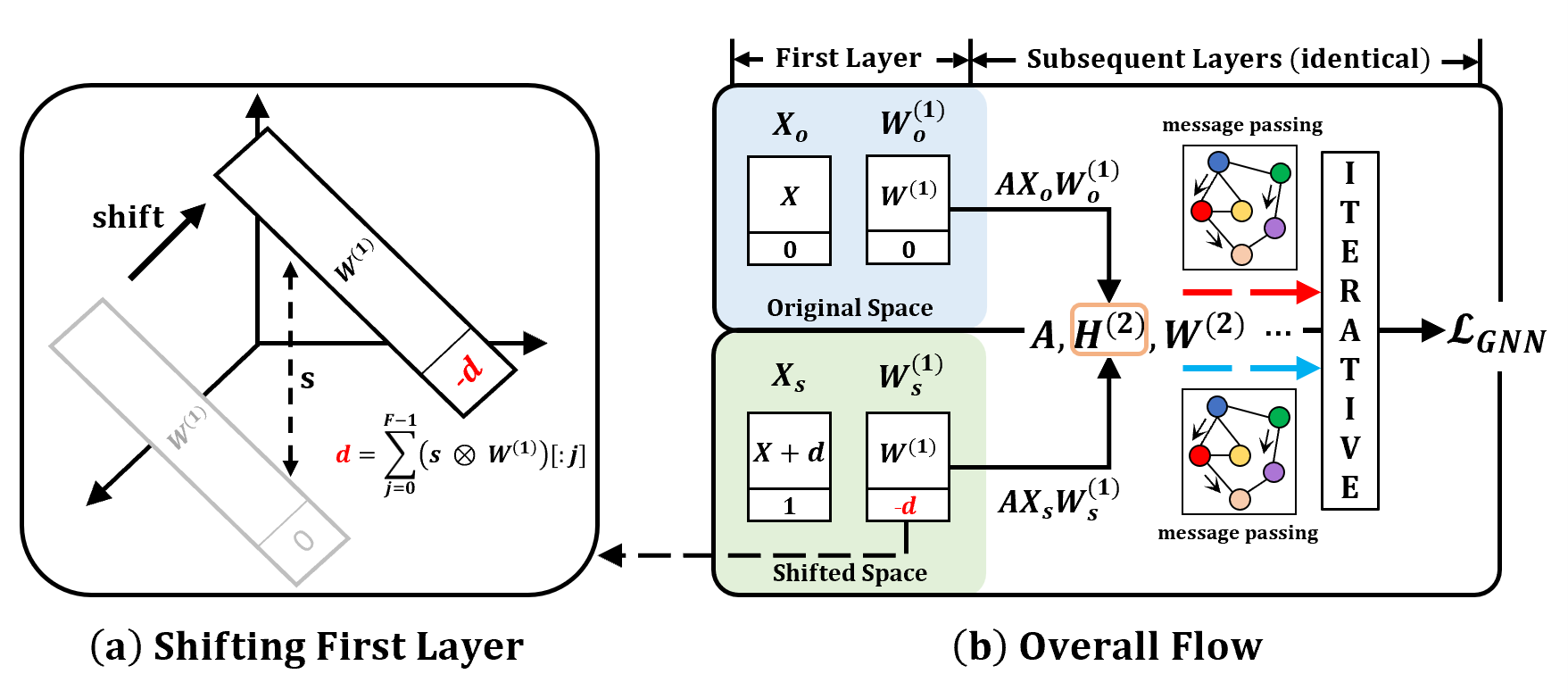}
    \caption{The figure above illustrates (a) the method of shifting the first matrix $W_1$ and (b) the overall architecture of Shift-GNN. As shown, the parameters are shared and updated iteratively in both spaces}
  \label{model}
\end{figure}

\section{METHODOLOGY} \label{methodology}
\subsection{Iterative Learning}
We introduce a strategy that simultaneously adjusts both feature vectors and the hyperplane. Straightforward shifting of the features can incur the regularization problem, depending on the magnitude of the added value (e.g., consider $\varepsilon=100$ in Eq. \ref{shifting} for clarity). To address this issue, we propose a method in which the first hyperplane is shifted alongside the feature (Fig \ref{model}a). We prove that the proposed scheme is more stable than flipping \cite{choi2022perturb}.
%(Detailed proof is in $\S$ \ref{theo_anal}). 
In Figure \ref{model}b, we illustrate a method for optimization through co-training in both the original (top) and shifted (bottom) spaces. Our approach utilizes the same parameter shared across both spaces, with the initial features $X$ and the first hyperplane $W^{(1)}$ being adjusted in each iteration. Since all parameters except for $W^{(1)}$ are identical in both spaces, we use the subscripts $W^{(1)}_o$ and $W^{(1)}_s$ to distinguish the two spaces. Additionally, the partial derivative of the loss function is represented by $\bigtriangledown$. The method for iteratively updating $W^{(1)}$ in both the original and shifted spaces is detailed below.

\textbf{Original space.} As shown in Figure \ref{model}b, we follow the mechanism of GCN \cite{kipf2016semi}, which takes $X_o$ and $W_o^{(1)}$ as inputs. One difference is that the initial feature matrix $X$ and the first hyperplane $W^{(1)}$ are zero-padded as shown below,
\begin{equation}
\begin{gathered}
\label{orig_plane}
X_o = \left(
    \begin{array}{cc}
      X\\
      0
    \end{array}
  \right), \,\,\, W_o^{(1)}=\left(
    \begin{array}{cc}
      W^{(1)}\\
      0
    \end{array}
  \right) .
\end{gathered}
\end{equation}
To maintain dimensional consistency, zero-padding is necessary since the last dimension is used in the shifted space for ease of implementation. Using $X_o$ and $W^{(1)}_o$, the output $\widehat{Y}$ can be derived through message-passsing (Eq. \ref{gnn}).

\textbf{Shifted space.} We first define a symbol $s \in \mathcal{R}^F$, which plays the role of shifting vector. Although various points could be used (e.g., the mean of all nodes), we choose a small $F$-dimensional vector $s=(0.1, \ldots, 0.1) \in \mathcal{R}^F$. The shifted feature $X_s$ is obtained by adding $s$ to $X$ and padding it with 1 as below:
\begin{equation}
\label{flip_init}
\begin{gathered}
X_s = \left(
    \begin{array}{c}
      X+s\\
      1
    \end{array}
  \right)
\end{gathered}
\end{equation}
To align with the shifted feature vectors, the first hyperplane $W^{(1)}$ must also be adjusted as illustrated in Figure \ref{model} a). To create $W^{(1)}_s$, we determine a distance vector $d \in \mathcal{R}^{F'}$ between the original hyperplane $W^{(1)}$ and a small value vector $s$ as follows:
\begin{equation}
\label{flip_plane}
\begin{gathered}
W^{(1)}_s=\left(
    \begin{array}{c}
      W^{(1)}\\
      - d
    \end{array}
  \right), \,\,\, d = \sum^{F-1}_{j=0} (s \otimes W^{(1)})[:j]
\end{gathered}
\end{equation}
The symbol $\otimes$ represents the element-wise product. Referring back to Figure \ref{model}a, we see that $W^{(1)}_s$ is obtained by adding $-d$ as the last element of $W^{(1)}$, ensuring that the outputs of both spaces are identical as $X_oW_o^{(1)} = X_sW^{(1)}_s$. This concurrent shifting of the hyperplane maintains the pairwise distance of hidden representations in both spaces (theoretical analysis is provided in $\S$ \ref{theo_anal}). Since the output is identical to the original space, we can compute the updated feature $H^{(2)}$ through a message-passing and an activation function $\sigma$ (e.g., ReLU) as follows:
\begin{equation}
\label{equality}
H^{(2)} = \sigma (AX_sW^{(1)}_s)
\end{equation}
Given that the equality $H_s^{(l)}=H_o^{(l)}$ holds ($l \geq 2$) after the first layer, the subsequent layers in both spaces can be derived as below:
\begin{equation}
H^{(l+1)}=\sigma(AH^{(l)}W^{(l)})
\end{equation}
Finally, the loss $\mathcal{L}_{GNN}$ is given by:
\begin{equation}
\label{loss_flip_gnn}
\mathcal{L}_{GNN}=\mathcal{L}_{nll}(Y, \widehat{Y})
\end{equation}
where $\widehat{Y}=softmax(H^{(L)})$. Using Eq. \ref{loss_flip_gnn}, the parameters of the GNNs can be updated iteratively in both spaces.

\subsection{Gradient and Convergence}
\textbf{Gradient.} It is worthwhile to note the following equation,
\begin{equation}
H^{(2)}=\sigma(AX_oW_o^{(1)})=\sigma(AX_sW_s^{(1)}), 
\end{equation}
which eventually implies that the gradients of two different spaces are equivalent for layers $l\geq2$. 
%after the first layer $(l\geq2)$ as below:
%\begin{equation}
%\bigtriangledown_{W_o^{(l)}}J=\bigtriangledown_{W_s^{(l)}}J 
%\end{equation}
%We assume that $J=\mathcal{L}_{GNN}$ is a full-batch loss function from both spaces. 
The only difference is in the gradient of the first layer.
%, which can be retrieved by referring to Eq. \ref{par_w1} as below:

\begin{table*}
\caption{Statistical details of nine benchmark graph datasets}
\label{dataset}
\centering
\begin{center}
\begin{adjustbox}{width=.7\textwidth}
\begin{tabular}{@{}ccccccccccc}
\multicolumn{1}{l}{}    & \multicolumn{1}{l}{}    &        &         &  & & & & & &\\ 
\Xhline{2\arrayrulewidth}
        & Datasets         & Cora  & Citeseer & Pubmed & Actor & Chameleon & Squirrel & Cornell & Texas & Wisconsin \\ 
\Xhline{2\arrayrulewidth}
                        & \# Nodes  & 2,708  & 3,327   & 19,717 & 7,600 & 2,277  & 5,201 & 183 & 183 & 251 \\
                        & \# Edges         & 10,558  & 9,104  & 88,648   & 25,944 & 33,824  & 211,872 & 295 & 309 & 499\\
                        & \# Features       & 1,433  & 3,703  & 500   & 931 & 2,325  & 2,089 & 1,703 & 1,703 & 1,703 \\
                        & \# Classes        & 7  & 6  & 3     & 5  & 5  & 5 & 5 & 5 & 5     \\
                        & \# Train      & 140  & 120  & 60  & 100  & 100  & 100 & 25 & 25 & 25\\
                        & \# Valid  & 500  & 500  & 500  & 3,750 & 1,088  & 2,550 & 79 & 79 & 113 \\
                        & \# Test           & 1,000  & 1,000  & 1,000  & 3,750 & 1,089  & 2,551 & 79 & 79 & 113     \\
\Xhline{2\arrayrulewidth}
\end{tabular}
\end{adjustbox}
\end{center}
\end{table*}

In the original and shifted spaces, $W_o^{(1)}$ and $W_s^{(1)}$ are updated as,
\begin{equation}
\label{orig_w1}
\begin{gathered}
\bigtriangledown_{W_o^{(1)}}J=(AX_o)^T\bigtriangledown_{\bar{H}^{(2)}}J ,
\end{gathered}
\end{equation}
and 
%Similarly, in the shifted space, $W_s^{(1)}$ follows:
\begin{equation}
\label{flip_w1}
\begin{gathered}
\bigtriangledown_{W_s^{(1)}}J=(AX_s)^T\bigtriangledown_{\bar{H}^{(2)}}J ,
\end{gathered}
\end{equation}
respectively.
In here,  $\bar{H}^{(2)}=AX_oW_o^{(1)}=AX_sW_s^{(1)}$. As demonstrated above, the main difference between the two equations lies in $AX_o$ and $AX_s$, which means that the gradient depends on the non-zero components in $X_o$ and $X_s$. Next, we prove that shifting ensures convergence and explain how to control the scale of gradients during back-propagation to ensure stable convergence.

\textbf{Convergence.} We aim to show that the optimization process guarantees the convergence of $W^{(1)}$. Excluding the padded (last) component in the feature vectors $X_o$ and $X_s$, we can reformulate Eqs. \ref{orig_w1} and \ref{flip_w1} as follows:
\begin{equation}
\begin{gathered}
\bigtriangledown_{W_o^{(1)}}J=A^TX^T\bigtriangledown_{\bar{H}^{(2)}}J, \,\,\,
\bigtriangledown_{W_s^{(1)}}J=A^T(X+s)^T\bigtriangledown_{\bar{H}^{(2)}}J
\end{gathered}
\end{equation}
The gradient of each dimension in $W^{(1)}$ is proportional to that of $X$ and $X + s$, respectively. As such, as training proceeds, $W_o$ converges to a local optimum $W^{(1)}_*$, $\lim\limits_{T \rightarrow \infty}\mathbb{E}[W^T_o-W^T_*]^2$, proportional to $\lim\limits_{T \rightarrow \infty}{\log{T} \over T}=0$. This is because two-layer neural networks with ReLU activation converge to a local minimum \cite{li2017convergence}, where the gradient $\bigtriangledown_{W_o} J$ and parameters $|W^{(l)}_o|$ are bounded. Similarly, $W_s$ is updated as $\lim\limits_{T \rightarrow \infty}\mathbb{E}[W^T_s-W^T_*]^2$, following $\lim\limits_{T \rightarrow \infty}{\log{(T+\varepsilon)} \over T}=0$. Compared to flip-based learning \cite{choi2022perturb}, a small shift value ($\varepsilon$) ensures more stable convergence.
Based on these properties, we can ensure the convergence of the parameters \cite{chen2017stochastic}. Detailed description of perturbed gradient descent can be found in \cite{jin2017escape}. Lastly, we adjust the scale of gradients in the shifted space to stabilize the model as, $W_s = W_s-z\bigtriangledown_{W_s}J$, where $z$ is the $z$ value of ego nodes.

\subsection{Theoretical Analysis of Performance Gains} \label{theo_anal}
In the context of empirical risk minimization, data augmentation plays a key role in the bias-variance tradeoff \cite{chen2020group}. Within this framework, we demonstrate that shifting functions as an effective augmentation strategy (\textbf{Thm.} \ref{thm_1}), enhancing the generalization of trainable parameters by reducing prediction variance (\textbf{Thm.} \ref{thm_2}).

\begin{manualtheorem}{1}[Shifting reduces the prediction variance]
    Consider the plane estimator \( g(X_o) = GNN(X_o) \), trained solely on the original features \( X_o \). Additionally, we introduce the augmented network \( \bar{g}(X) = GNN(X) \), which utilizes both feature sets \( X = \{X_o \cup X_s\} \). It is clear that the function \( g \) is invariant to shifting, as \( g(X_o, W_o) = g(X_s, W_s) \), with \( X_s \) maintaining the pairwise distances between nodes. Given that the bias term remains invariant (due to the concurrent shift of a hyperplane), our focus turns to the variance of \( g(X) \), which can be decomposed using the law of total variance:
    \begin{equation}
    \label{variance}
    \begin{gathered}
    \mathbb{V}[g(X)] = \mathbb{V}[\bar{g}(X)] + \mathbb{E}\left[\mathbb{V}[ g(X)]\right]
    \end{gathered}
    \end{equation}
    In the above equation, we can infer that $\mathbb{V} [\mathbb{E}\left[g(X)\right]]=\mathbb{V}[\bar{g}(X)]$ since they share the same marginal distribution. Also, the Wasserstein distance (e.g., $L_2$) between the two distributions $\mathcal{W}_1(\mathbb{E}\left[g(X)\right],\mathbb{E}\left[\bar{g}(X)\right])$ (the difference in their means) is independent of the total variance. From this observation, we can establish the following conditions:
    \begin{equation}
    \label{covariance}
    \mathbb{V}[\bar{g}(X)] \leq \mathbb{V}[g(X)]
    \end{equation}
    %Finally, we show that the difference between two networks can be derived as below:
    %\begin{equation}
    %\label{perf_gain}
    %\mathbb{V}[\bar{g}(X)]-\mathbb{V}[g(X)] \in -\mathbb{E}\left[tr(\mathbb{V}[g(X)]\right]
    %\end{equation}
    This suggests that the performance improvement of the augmented model is related to a reduction in variance compared to the baseline method. In addition, we can prove that variance reduction improves classification accuracy. We can prove this through the well-known bias-variance tradeoff as below. 
\label{thm_1}
\end{manualtheorem}

\begin{table*}
\caption{(RQ1) We denote the performance with the highest node classification accuracy in bold ($^\ast$) on nine benchmark datasets}
\label{perf_1}
\centering
\begin{center}
\begin{adjustbox}{width=\textwidth}
\begin{tabular}{lllllllllll}
&  & & &              \\ 
\Xhline{2\arrayrulewidth}
\Xhline{2\arrayrulewidth}
        & Datasets                       & Cora  & Citeseer & Pubmed & Actor & Cornell & Texas & Wisconsin & Chameleon & Squirrel \\ 
\Xhline{2\arrayrulewidth}
        & $z$-value  & 0.59 & 0.41 & 0.96 & 0.21 & 0.62 & 0.41 & 0.53 & 1.0 & 1.0 \\
        & Homophily \cite{pei2020geom} & 0.81 & 0.74 & 0.8 & 0.22 & 0.11 & 0.06 & 0.16 & 0.23 & 0.22 \\
\Xhline{2\arrayrulewidth}
                        & MLP \cite{popescu2009multilayer}             & 53.2 $_{\,\pm\,0.5\,\%}$ & 53.7 $_{\,\pm\,1.7\,\%}$ & 69.7 $_{\,\pm\,0.4\,\%}$ & 27.9 $_{\,\pm\,1.1\,\%}$ & 60.1 $_{\,\pm\,1.2\,\%}$ & 65.8 $_{\,\pm\,5.0\,\%}$ &  73.5 $_{\,\pm\,5.4\,\%}$ & 41.2 $_{\,\pm\,1.8\,\%}$ & 26.5 $_{\,\pm\,0.6\,\%}$\\
                        & GCN \cite{kipf2016semi}              & 79.1 $_{\,\pm\,0.7\,\%}$  & 67.5 $_{\,\pm\,0.3\,\%}$ & 77.8 $_{\,\pm\,0.2\,\%}$ & 20.4 $_{\,\pm\,0.6\,\%}$  & 39.4 $_{\,\pm\,4.3\,\%}$ & 47.6 $_{\,\pm\,0.7\,\%}$ & 40.5 $_{\,\pm\,1.9\,\%}$ & 49.4 $_{\,\pm\,0.7\,\%}$ & 31.8 $_{\,\pm\,0.9\,\%}$ \\
                        & Ortho-GCN \cite{guo2022orthogonal}      & 80.6 $_{\,\pm\,0.4\,\%}$ & 69.5 $_{\,\pm\,0.3\,\%}$ & 76.9 $_{\,\pm\,0.3\,\%}$ & 21.4 $_{\,\pm\,1.6\,\%}$ & 45.4 $_{\,\pm\,4.7\,\%}$ & 53.1 $_{\,\pm\,3.9\,\%}$ & 46.6 $_{\,\pm\,5.8\,\%}$ & 46.7 $_{\,\pm\,0.5\,\%}$ & 31.3 $_{\,\pm\,0.6\,\%}$ \\
                        & GAT \cite{velickovic2017graph}            & 80.1 $_{\,\pm\,0.6\,\%}$ & 68.0 $_{\,\pm\,0.7\,\%}$ & 78.0 $_{\,\pm\,0.4\,\%}$ & 22.5 $_{\,\pm\,0.3\,\%}$ & 42.1 $_{\,\pm\,3.1\,\%}$ & 49.2 $_{\,\pm\,4.4\,\%}$ & 45.8 $_{\,\pm\,5.3\,\%}$ & 46.9 $_{\,\pm\,0.8\,\%}$ & 30.8 $_{\,\pm\,0.9\,\%}$ \\
                        & APPNP \cite{klicpera2018predict}                & 81.2 $_{\,\pm\,0.4\,\%}$ & 68.9 $_{\,\pm\,0.3\,\%}$ & 79.0 $_{\,\pm\,0.4\,\%}$ & 21.5 $_{\,\pm\,0.2\,\%}$ & 49.8 $_{\,\pm\,3.6\,\%}$ & 56.1 $_{\,\pm\,0.2\,\%}$ & 45.7 $_{\,\pm\,1.7\,\%}$ & 45.0 $_{\,\pm\,0.5\,\%}$ & 30.3 $_{\,\pm\,0.6\,\%}$\\
                        & GIN \cite{xu2018powerful}             & 77.3 $_{\,\pm\,0.8\,\%}$ & 66.1 $_{\,\pm\,0.6\,\%}$ & 77.1 $_{\,\pm\,0.7\,\%}$ & 24.6 $_{\,\pm\,0.8\,\%}$ & 42.9 $_{\,\pm\,4.6\,\%}$ & 53.5 $_{\,\pm\,3.0\,\%}$ & 38.2 $_{\,\pm\,1.5\,\%}$ & 49.1 $_{\,\pm\,0.7\,\%}$ & 28.4 $_{\,\pm\,2.2\,\%}$\\
                        & GCNII \cite{chen2020simple}          & 80.8 $_{\,\pm\,0.7\,\%}$ & 69.0 $_{\,\pm\,1.4\,\%}$ & 78.8 $_{\,\pm\,0.4\,\%}$ & 26.1 $_{\,\pm\,1.2\,\%}$ &  62.5 $_{\,\pm\,0.5\,\%}$ &  69.3 $_{\,\pm\,2.1\,\%}$ & 63.2 $_{\,\pm\,3.0\,\%}$ & 45.1 $_{\,\pm\,0.5\,\%}$ & 28.1 $_{\,\pm\,0.7\,\%}$ \\
                        & H$_2$GCN \cite{zhu2020beyond}       & 79.5 $_{\,\pm\,0.6\,\%}$ & 67.4 $_{\,\pm\,0.5\,\%}$ & 78.7 $_{\,\pm\,0.3\,\%}$ & 25.8 $_{\,\pm\,1.2\,\%}$ & 59.8 $_{\,\pm\,3.7\,\%}$ & 66.3 $_{\,\pm\,4.6\,\%}$ & 61.5 $_{\,\pm\,4.4\,\%}$ & 47.3 $_{\,\pm\,0.9\,\%}$ & 31.1 $_{\,\pm\,0.5\,\%}$ \\
                        & FAGCN \cite{bo2021beyond}       & 81.0 $_{\,\pm\,0.3\,\%}$ & 68.3 $_{\,\pm\,0.6\,\%}$ & 78.9 $_{\,\pm\,0.4\,\%}$ & 26.7 $_{\,\pm\,0.8\,\%}$ & 46.5 $_{\,\pm\,1.7\,\%}$ & 53.8 $_{\,\pm\,1.2\,\%}$ & 51.0 $_{\,\pm\,4.1\,\%}$ & 46.8 $_{\,\pm\,0.6\,\%}$ & 29.9 $_{\,\pm\,0.5\,\%}$ \\
                        & GPR-GNN \cite{chien2020adaptive} & 82.2 $_{\,\pm\,0.4\,\%}$ & 70.1 $_{\,\pm\,0.8\,\%}$ &  79.3 $_{\,\pm\,0.3\,\%}$ & 25.1 $_{\,\pm\,0.5\,\%}$ & 53.1 $_{\,\pm\,1.6\,\%}$ & 61.2 $_{\,\pm\,0.9\,\%}$ & 62.4 $_{\,\pm\,1.2\,\%}$ & 50.9 $_{\,\pm\,0.8\,\%}$ & 30.4 $_{\,\pm\,0.4\,\%}$ \\
                        & ACM-GCN \cite{luan2022revisiting} & 80.2 $_{\,\pm\,0.8\,\%}$ & 68.3 $_{\,\pm\,1.1\,\%}$ & 78.1 $_{\,\pm\,0.5\,\%}$ & 24.9 $_{\,\pm\,2.0\,\%}$ & 55.6 $_{\,\pm\,3.3\,\%}$ & 58.9 $_{\,\pm\,2.6\,\%}$ & 61.3 $_{\,\pm\,0.5\,\%}$ & 49.5 $_{\,\pm\,0.7\,\%}$ & 31.6 $_{\,\pm\,0.4\,\%}$ \\
                        & JacobiConv \cite{wang2022powerful} & 81.9 $_{\,\pm\,0.6\,\%}$ & 69.6 $_{\,\pm\,0.8\,\%}$ & 78.5 $_{\,\pm\,0.4\,\%}$ & 25.7 $_{\,\pm\,1.2\,\%}$ & 55.3 $_{\,\pm\,3.4\,\%}$ & 57.7 $_{\,\pm\,3.6\,\%}$ & 53.4 $_{\,\pm\,1.6\,\%}$ &  \textbf{52.8$^*$ $_{\,\pm\,0.9\,\%}$} & 32.0 $_{\,\pm\,0.6\,\%}$ \\
                        & AERO-GNN \cite{lee2023towards} & 81.6 $_{\,\pm\,0.5\,\% }$ & 71.1 $_{\,\pm\,0.6\,\% }$ & 79.1 $_{\,\pm\,0.4\,\% }$ & 25.5 $_{\,\pm\,1.0\,\% }$ & 47.7 $_{\,\pm\,4.2\,\% }$ & 50.4 $_{\,\pm\,3.5\,\% }$ & 42.6 $_{\,\pm\,2.2\,\% }$ & 49.8 $_{\,\pm\,2.3\,\% }$ & 29.9 $_{\,\pm\,1.9\,\% }$ \\
                        & Auto-HeG \cite{zheng2023auto} & 81.5 $_{\,\pm\,1.1\,\% }$ & 70.9 $_{\,\pm\,1.4\,\% }$ & 79.2 $_{\,\pm\,0.2\,\% }$ & 26.1 $_{\,\pm\,0.9\,\% }$ & 55.5 $_{\,\pm\,3.5\,\% }$ & 60.2 $_{\,\pm\,3.0\,\% }$ & 62.3 $_{\,\pm\,2.7\,\% }$ & 48.7 $_{\,\pm\,1.3\,\% }$ & 31.5 $_{\,\pm\,1.1\,\% }$\\
                        & TED-GCN \cite{yan2024trainable} & 81.8 $_{\,\pm\,0.8\,\% }$ & 71.4 $_{\,\pm\,0.5\,\% }$ & 78.6 $_{\,\pm\,0.3\,\% }$ & 26.0 $_{\,\pm\,0.9\,\% }$ & 54.6 $_{\,\pm\,3.2\,\% }$ & 59.0 $_{\,\pm\,4.8\,\% }$ & 61.7 $_{\,\pm\,2.6\,\% }$ & 50.4 $_{\,\pm\,1.2\,\% }$ & \textbf{33.0}$^*$ $_{\,\pm\,0.9\,\% }$ \\
                        & PCNet \cite{li2024pc} & 81.5 $_{\,\pm\,0.7\,\% }$ & 71.2 $_{\,\pm\,1.2\,\% }$ & 78.8 $_{\,\pm\,0.2\,\% }$ & 26.4 $_{\,\pm\,0.8\,\% }$ & 55.0 $_{\,\pm\,3.1\,\% }$ & 60.2 $_{\,\pm\,2.4\,\% }$ & 63.0 $_{\,\pm\,1.8\,\% }$ & 48.1 $_{\,\pm\,1.6\,\% }$ & 31.4 $_{\,\pm\,0.5\,\% }$ \\
\Xhline{2\arrayrulewidth}
                        & Shift-MLP      & 61.8 $_{\,\pm\,0.5\,\%}$ & 60.4 $_{\,\pm\,0.5\,\%}$ & 74.4 $_{\,\pm\,0.6\,\%}$ &  \textbf{36.0$^*$ $_{\,\pm\,0.4\,\%}$} &  \textbf{72.3$^*$ $_{\,\pm\,3.5\,\%}$} &  \textbf{81.0$^*$ $_{\,\pm\,2.5\,\%}$} &  \textbf{80.9$^*$ $_{\,\pm\,4.2\,\%}$} & 43.1 $_{\,\pm\,0.9\,\%}$ & 27.5 $_{\,\pm\,0.6\,\%}$ \\
                        & vs MLP (+ \%) & + 16.2 \% & + 12.5 \% & + 6.7 \% & \textbf{+ 29.0} \% & \textbf{+ 20.3} \% & \textbf{+ 23.1} \% & \textbf{+ 10.1} \% & + 4.6 \% & + 3.8 \% \\
\Xhline{2\arrayrulewidth}
                        & Shift-GCN      & 82.9 $_{\,\pm\,0.4\,\%}$ & 72.6 $_{\,\pm\,0.4\,\%}$ & 79.2 $_{\,\pm\,0.2\,\%}$ & 29.1 $_{\,\pm\,0.3\,\%}$ & 49.4 $_{\,\pm\,0.5\,\%}$ & 64.2 $_{\,\pm\,1.4\,\%}$ & 52.6 $_{\,\pm\,2.1\,\%}$ & 51.0 $_{\,\pm\,0.5\,\%}$ & 32.1 $_{\,\pm\,0.3\,\%}$\\
                        & vs GCN (+ \%) & + 4.7 \% & + 7.7 \% & + 1.5 \% & + 42.6 \% & + 25.4 \% & + 35.3 \% & + 29.9 \% & + 2.1 \% & + 2.0 \% \\
\Xhline{2\arrayrulewidth}
                        & Shift-GAT        &  \textbf{83.3} $^*_{\,\pm\,0.4\,\%}$ &  \textbf{73.0} $^*_{\,\pm\,0.3\,\%}$ & 79.0 $_{\,\pm\,0.2\,\%}$ & 30.6 $_{\,\pm\,0.6\,\%}$ & 52.1 $_{\,\pm\,1.1\,\%}$ & 61.4 $_{\,\pm\,0.8\,\%}$ & 54.5 $_{\,\pm\,1.7\,\%}$ & 48.3 $_{\,\pm\,0.3\,\%}$ & 32.5 $_{\,\pm\,0.4\,\%}$ \\ 
                        & vs GAT (+ \%) & + \textbf{4.0 \%} & \textbf{+ 7.4} \% & + 1.3 \% & + 36.0 \% & + 19.2 \% & + 24.8 \% & + 19.0 \% & + 3.0 \% & + 3.6 \% \\
\Xhline{2\arrayrulewidth}
                        & Shift-FAGCN        &  82.9 $_{\,\pm\,0.4\,\%}$ &  72.5 $_{\,\pm\,1.4\,\%}$ &  \textbf{79.6$^*$ $_{\,\pm\,0.3\,\%}$} &  34.7 $_{\,\pm\,0.6\,\%}$ & 54.0 $_{\,\pm\,1.8\,\%}$ & 60.1 $_{\,\pm\,0.9\,\%}$ & 64.6 $_{\,\pm\,2.7\,\%}$ & 48.1 $_{\,\pm\,0.4\,\%}$ & 30.3 $_{\,\pm\,0.4\,\%}$ \\               
                        & vs FAGCN (+ \%)  & +2.3 \% & +6.1 \% & \textbf{+1.0 \%} & +30.0 \% & +16.1 \% & +11.7 \% & +26.7 \% & +1.1 \% & +0.6 \% \\
\Xhline{2\arrayrulewidth}                        
\Xhline{2\arrayrulewidth}
\end{tabular}
\end{adjustbox}
\end{center}
\end{table*}

\begin{manualtheorem}{2}[Bias-variance tradeoff in cross-entropy]
    Like the previous theorem, let us consider an input $X$ and its corresponding output $g(X)$ with the true label $Y$. Then, the cross-entropy error can be decomposed using the Kullback-Leibler divergence ($D_{KL}$) \cite{yang2020rethinking} as follows:
    \begin{equation}
        Err(X)=\underset{\text{bias}^2}{\underline{D_{KL}(Y \, || \, \mathbb{E}[g(X)])}}+\underset{\text{variance}}{\underline{\mathbb{E}[D_{KL}(\mathbb{E}[g(X)] \, || \, g(X))]}}+\eta^2
    \end{equation}
    The first term represents the squared bias, while the second term corresponds to the variance. The symbol $\eta$ denotes the irreducible error. Note that the bias (first term) remains unchanged during concurrent shifting. In contrast, the variance (second term) of the shifted model, given by $\mathbb{V}[\bar{g}(X)]$ in Eq. \ref{covariance}, decreases compared to that of the original model, $\mathbb{V}[g(X)]$. This result can be derived using the \textit{Loewner order} \cite{chen2020group} as follows:
    \begin{equation}
        \mathbb{V}[\bar{g}(X)] - \mathbb{V}[g(X)] \approx -\mathbb{E}\big[\mathbb{V}[\bar{g}(X)]\big] \pm \Delta
    \end{equation}
    where $\Delta=4||g||_{\infty} \mathbb{E}\big[\mathcal{W}_1(\bar{g}(X),g(X))]$. Thus, we can conclude that a reduction in variance enhances classification performance.
\label{thm_2}
\end{manualtheorem}

\subsection{Time Complexity}
Given the GCN \cite{kipf2016semi}, we can segment our model into the original and the shifted spaces. The computational load for the GCN is established as $\mathcal{O}(|\mathcal{E}|P_{GCN})$, with $P_{GCN}$ representing the total count of learnable parameters. More precisely, $P_{GCN}$ is further detailed \cite{zhu2020beyond} as $\mathcal{O}(nz(X)F'+F'C)$, where $nz(X)$ signifies the quantity of non-zero elements in the input matrix $X$, $F'$ denotes the dimensions post-projection, and $F'C$ refers to the parameters within the secondary convolutional layer. %Furthermore, the notation $\mathcal{O}|\mathcal{E}|\approx\mathcal{O}|\mathcal{E}|d_{max}$ is utilized, with $d_{max}$ indicating the maximum node degree. 
Mirroring the baseline GCN, the computational complexity of the shifted GCN is expressed as $\mathcal{O}(|\mathcal{E}|P'_{GCN})$. Specifically, $P'_{GCN}=\mathcal{O}(nz(X_s)F'+F'C)$, herein $nz(X_s)$ counts the non-zero components within the shifted features. Upon synthesizing these computations, the overall complexity of our approach is articulated as $\mathcal{O}(|\mathcal{E}|(P_{GCN}+P'_{GCN}))$, directly correlating with the edge count and the cumulative dimensions of the adjustable matrices.

\section{EXPERIMENTS} \label{experiments}
We conduct experiments to answer the following questions:
\begin{itemize}
\item \textbf{RQ1:} Does shifting effectively mitigate the problem caused by sparse initial features?
\item \textbf{RQ2:} How does our shifting technique, which combines feature perturbation with hyperplane adjustment, improve performance compared to feature perturbation only?
\item \textbf{RQ3:} Does the proposed shifting method outperform existing techniques for sparse modeling?
\item \textbf{RQ4:} How significant is the difference between the gradients from the original and shifted spaces?
%\item \textbf{RQ5:} To what extent does the choice of hyperparameters, as specified in Equation \ref{hyperparam}, influence overall outcomes?
\item \textbf{RQ5:} What are the performance implications of applying shifting exclusively to features or the hyperplane?
\end{itemize}
%Due to the limited space, please find more experimental results in our full paper \cite{choi2022perturb}.

\subsection{Datasets, Baselines, and Implementation} \label{data_desc}
%The datasets, baselines, and implementations are as follows.

\textbf{Datasets.} We perform our experiments on nine benchmark datasets shown in Table \ref{dataset}. The homophily in Table \ref{perf_1} is defined in \cite{pei2020geom}.
%\begin{equation}
%\label{homophily}
%h = {\sum_{(i,j)\in \mathcal{E}} 1(Y_i=Y_j) \over |\mathcal{E}|},
%\end{equation}
%which represents the proportion of edges that connect two nodes with the same label. 
We employ the datasets frequently used in prior studies; (1) \textit{Cora, Citeseer, Pubmed} \cite{kipf2016semi}, (2) \textit{Actor} \cite{tang2009social}, (3) \textit{Chameleon, Squirrel} \cite{rozemberczki2019gemsec}, (4) \textit{Cornell, Texas,Wisconsin\footnote{http://www.cs.cmu.edu/~webkb/}}.

\textbf{Baselines.} We compare the performance of the proposed scheme with those of classical and state-of-the-art methods. The baseline models are as follows:
\begin{itemize}
    \item \textbf{MLP} \cite{popescu2009multilayer} adopts a feed-forward neural network without message-passing.
    \item \textbf{GCN} \cite{kipf2016semi} suggests the first-order approximation of Chebyshev polynomials \cite{defferrard2016convolutional} to preserve low frequency signals.
    %\item \textbf{DropEdge} \cite{rong2019dropedge} randomly deletes edges based on a given probability in order to alleviate over-fitting.
    \item \textbf{Ortho-GCN} \cite{guo2022orthogonal} maintains the orthogonality of feature transformation matrices through three constraints. 
    \item \textbf{GAT} \cite{velickovic2017graph} employs node features to assign weights for each edge without graph structural property.
    %\item \textbf{GATv2} \cite{brody2021attentive} improves GAT to generate a more dynamic graph attention that is more expressive.
    \item \textbf{APPNP} \cite{klicpera2018predict} combines personalized PageRank with GCN while reducing the computational cost.
    \item \textbf{GIN} \cite{xu2018powerful}
    adopts an injective mapping function to enhance the discriminative power of GCN and ensures isomorphism.
    \item \textbf{GCNII} \cite{chen2020simple} further utilizes the weighted identity mapping function to redeem the deficiency of APPNP. 
    \item \textbf{H$_2$GCN} \cite{zhu2020beyond} separates ego from neighbors and applies hop-based aggregation for heterophilic networks.
    %\item \textbf{P-reg} \cite{yang2021rethinking} improves the traditional graph Laplacian to provide extra information that GNNs might not capture.
    \item \textbf{FAGCN} \cite{bo2021beyond} adaptively controls the propagation of low and high-frequency signals during message-passing.
    \item \textbf{GPRGNN} \cite{chien2020adaptive} introduces an adaptive PageRank to generalize for both homophilic and heterophilic graphs.
    \item \textbf{ACM-GCN} \cite{luan2022revisiting} proposes an adaptive channel mixing to capture the diverse information of homophily and heterophily.
    \item \textbf{JacobiConv} \cite{wang2022powerful} analyzes the expressive power of spectral GNN and suggests a Jacobi basis convolution.
    %\item \textbf{GloGNN} \cite{li2022finding} generates the node embedding by receiving information from global nodes.
    \item \textbf{AERO-GNN} \cite{lee2023towards} improves the deep graph attention to reduce the smoothing effect at deep layers.
    \item \textbf{Auto-HeG} \cite{zheng2023auto} automatically build heterophilic GNN models with search space design and architecture selection. 
    \item \textbf{TED-GCN} \cite{yan2024trainable} redefines GCN’s depth $L$ as a trainable parameter, which can control its signal processing capability to model both homophily/heterophily graphs.
    \item \textbf{PCNet} \cite{li2024pc} proposes a two-fold filtering mechanism to extract homophily in heterophilic graphs.
\end{itemize}

\textbf{Implementation.} To begin with, we set the shifting value in Eq. \ref{flip_init} as $s=0.1$. All methods are implemented using \textit{PyTorch Geometric}\footnote{\label{footnote}https://pytorch-geometric.readthedocs.io/en/latest/modules/nn.html}. The hidden dimension was set to 64, with ReLU activation and dropout applied. Log Softmax was used for classification. The model was trained with a learning rate of $1e^{-3}$ and the Adam optimizer, including a weight decay of $5e^{-4}$. For the training set, we randomly selected 20 nodes per class, except for the WebKB dataset (5 for each class) due to its limited size. The remaining nodes were randomly split equally into validation and test sets, while we followed the procedure outlined in \cite{kipf2016semi} for the citation networks. Test accuracy with the best validation score was selected for comparison. Since the first layer of Shift-MLP and Shift-GCN is a simple projection matrix, we sequentially apply feature shifting, hyperplane shifting, and sign shifting. In contrast, Shift-GAT and Shift-FAGCN utilize an attention layer. However, the attention is computed after the projection layer, ensuring that the values remain identical in both spaces. Thus, the shifting technique for the initial projection matrix can also be applied to other state-of-the-art methods \cite{chien2020adaptive,luan2022revisiting,wang2022powerful}.

\subsection{(RQ1) Node Classification Performance}
We first show that the performance gain is sensitive to $z$ as shown in Table \ref{perf_1}. Since shifting is designed to mitigate overfitting caused by sparse initial features, it is reasonable to assume that the value $z$ is a key factor determining the performance gain. Indeed, shifting achieves greater performance gains on datasets with low $z$ values (i.e. sparse initial features) compared to those with high $z$ values (dense initial features). For three datasets with large $z$ (Pubmed, Chameleon, and Squirrel), the improvement of shifting over their vanilla models (e.g., Shift-MLP vs. MLP) is relatively modest. Nonetheless, four shifting variants outperform the original methods, achieving relative improvements of 4.8\%, 1.9\%, 2.6\%, and 0.9\%, respectively, indicating that shifting can is effective for the datasets with dense initial features. On another set of graphs with low $z$ (six datasets), the shifted methods substantially improve their base models, with the greater average gains of 18.5\%, 24.2\%, 18.4\%, and 15.5\%. Notably, Shift-based algorithms show the best performance, except for Chameleon and Squirrel (which have high $z$ and low homophily), implying that input feature perturbation can have a greater impact than aggregation schemes.

In Table \ref{perf_1}, we can observe that performance of the shifting mechanism is influenced by the homophily ratio. Message-passing GNNs leverage the homophily property commonly observed in graphs \cite{lim2021large,yan2021two}. Among the three citation graphs (Cora, Citeseer, and Pubmed), GNNs outperform Multi-Layer Perceptrons (MLPs) due to their higher homophily ratios. However, in other datasets, such as Actor and the three WebKB networks, MLP achieves the highest accuracy among the baselines, indicating that message-passing fails to generalize effectively in the presence of high heterophily. While the performance gain of Shift-MLP is higher than Shift-GNNs on homophilic graphs, GNNs benefit more from shifting on heterophilic graphs. Although our shifting methods achieve marginal improvements on Chameleon and Squirrel, even when $z$ is extremely high, the reduction in variance accounts for the observed performance improvement as demonstrated in Theorem \ref{thm_2}. Instead, they exhibit substantial performance gains across seven graph datasets compared to state-of-the-art methods (GPR-GNN, AERO-GNN, JacobiConv, PCNet).

\begin{figure}[t]
\centering
    \includegraphics[width=.49\textwidth]{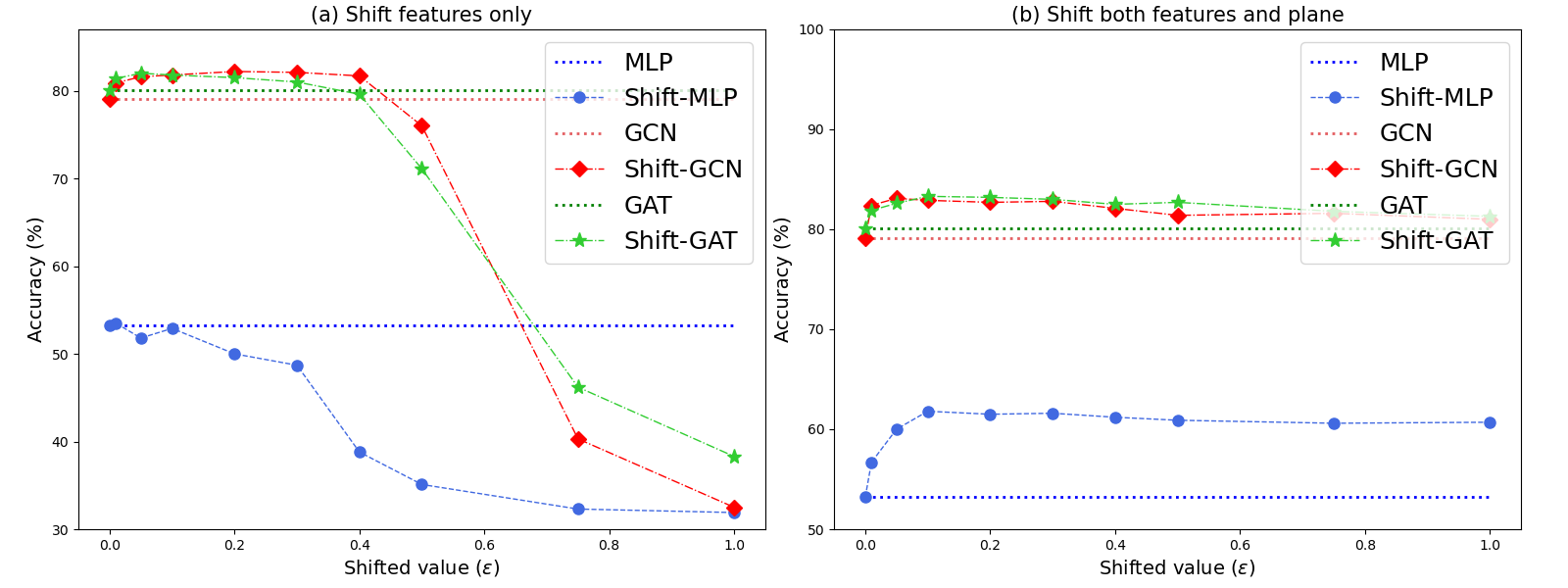}
    \caption{(RQ2) Node classification accuracy of MLP, GCN, GAT, and their shifted versions as functions of the parameter $\varepsilon$ on the Cora dataset. In the left figure, we shift only the features, whereas the right figure shows the performance when both the features and the first plane are shifted}
  \label{convergence}
\end{figure}

\subsection{(RQ2) Effect of Hyperplane Shifting}
We point out the drawback of simple initial feature vector perturbation and propose a scheme that integrates feature perturbation with the shifting of the first hyperplane. Figure \ref{convergence} compares the node classification accuracy of MLP, GCN, GAT, and their shifted versions as the shifted parameter $\varepsilon$ (Eq. \ref{shifting}) varies on the Cora dataset. Figure \ref{convergence} a) shows the results when only the features are disturbed. In this scenario, the accuracy of all shifted models decreases sharply as $\varepsilon$ increases, indicating that these models are sensitive to perturbations. Figure \ref{convergence} b) presents the performance when both the features and the first hyperplane are shifted simultaneously. Here, all shifted versions maintain stable accuracy even for large $\varepsilon$, demonstrating high robustness to the introduced shifts. This stability can be attributed to the concurrent shifting strategy, which preserves the output scale and mitigates issues such as gradient explosion or vanishing gradients. Furthermore, the findings suggest that adopting a concurrent shifting mechanism can effectively stabilize the performance of models under perturbative conditions, providing a practical approach to enhance model resilience.

\begin{table}[t]
\caption{(RQ3) The node classification accuracy of Shift-MLP and four sparse feature learning algorithms on two homophilic (Core, Citeseer) and two heterophilic (Cornell and Texas) datasets.}
\centering
\begin{adjustbox}{width=.45\textwidth}
\begin{tabular}{@{}lccccc}
&  & & &              \\ 
\Xhline{2\arrayrulewidth}
        & Datasets                       & Cora  & Citeseer & Cornell & Texas  \\ 
%\Xhline{2\arrayrulewidth}
%        & $z$ (Eq. \ref{nz_ratio})  & 0.59 & 0.41 & 0.96 & 1.0 \\
%        & Homophily (Eq. \ref{homophily}) & 0.81 & 0.74 & 0.8 & 0.22 \\
\Xhline{2\arrayrulewidth}
                        & SVD \cite{golub1971singular} & 52.1 & 44.5 & 46.7 & 51.0 \\
                        & Sparse PCA \cite{wold1987principal}              & 50.9 & 49.8 & 53.5 & 53.3 \\
                        & SF \cite{ngiam2011sparse}              & 59.2 & 52.3 & 64.0 & 64.2 \\
                        & MO-SFL \cite{gong2015multiobjective} & 54.5 & 50.6 & 63.9 & 66.5 \\
                        \Xhline{2\arrayrulewidth}
                        & Shift-MLP  & \textbf{61.8}$^*$ & \textbf{60.4}$^*$ & \textbf{72.3}$^*$ & \textbf{81.0}$^*$ \\
\Xhline{2\arrayrulewidth}
\label{sparse_perf}
\end{tabular}
\end{adjustbox}
\end{table}

\subsection{(RQ3) Sparse Modeling}
Table \ref{sparse_perf} presents the node classification accuracy of sparse modeling methods that are efficient for filtering a few non-zero elements. For a fair comparison, we employ Shift-MLP to exclude the message-passing scenario. Assume the initial node feature matrix has dimensions $X=\mathcal{R}^{n \times d}$. The baseline methods are implemented as follows: the latent dimension for both SVD \cite{golub1971singular} and PCA \cite{wold1987principal} is set to $d/2$. For SVD, $X=U\sum V^T$, and we use the output $V^T$ as the input to the MLP instead of $X$. For PCA, the top $d/2$ eigenvectors are selected for dimensionality reduction, and the reduced vector is used as the MLP input. However, the complexity of SVD \cite{golub1971singular} and PCA \cite{wold1987principal} is known to be $O(mn + n^3)$, making them potentially unsuitable for large graphs. SF \cite{ngiam2011sparse} is simple and intuitive, normalizing the row as $\bar{X}_j=X_j/||X_j||_2$ and the column as $\widehat{X}_i=\bar{X}_j/||\bar{X}||_2$. Lastly, we construct a reconstruction network for MO-SFL \cite{gong2015multiobjective}, using the latent representation as the input to the MLP. Although the other two methods have lower computational costs, SF \cite{ngiam2011sparse} fails to preserve the variance (Eq. \ref{variance}) of features, and MO-SFL \cite{gong2015multiobjective} requires an additional reconstruction step. In contrast, Shift-MLP performs best at a low computational cost. This result implies that our dual shifting scheme can be combined with generic neural networks and improves their performance.

\begin{figure}[t]
    \includegraphics[width=.47\textwidth]{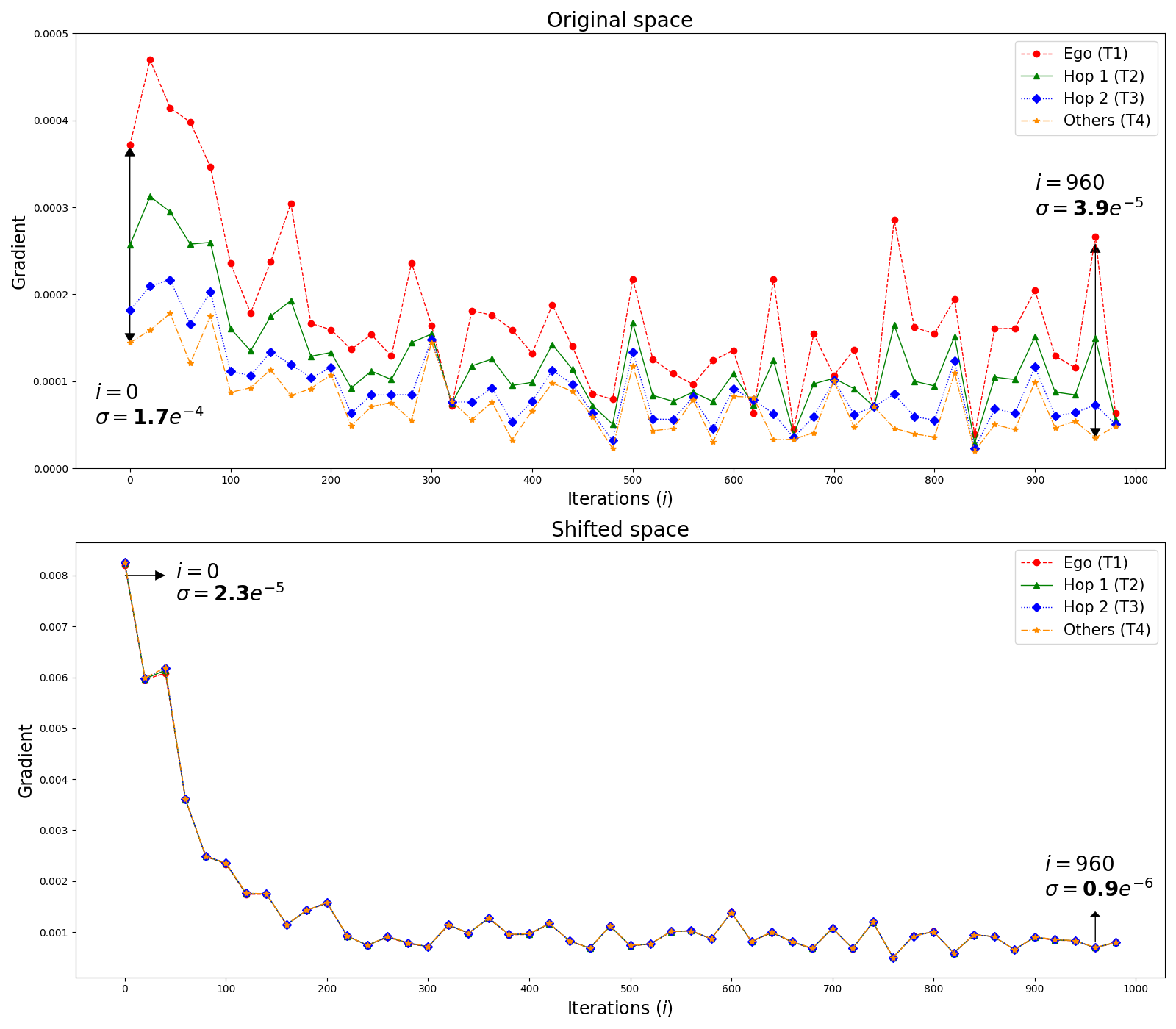}
    \caption{(RQ4) Using the Cora dataset, we plot the magnitude of the first projection matrix gradients and their standard deviation ($\sigma$) during training epochs ($i$)}
  \label{gradient}
\end{figure}

\subsection{(RQ4) Gradient Analysis} 
In Figure \ref{gradient}, we analyze the gradient of the first projection matrix during the training phase using the Cora dataset. We apply different ranges of neighboring nodes and define four types based on the range. Type 1 (T1) consists of the central node (Ego) only, while T2 and T3 include the features of 1-hop and 2-hop neighbors, respectively. T4 comprises the remaining features. To avoid overlap and duplicated attention of features, we prioritize the types from T1 to T4 (all $T \in \mathcal{R}^{F}$ are binarized vectors). The figure illustrates the average and standard deviation of gradients in original and shifted spaces, represented by different colors. In Figure \ref{gradient} a), which shows the gradients in the original space, the largest gradient is assigned to the features of central nodes, T1 (red) \cite{kang2022rawlsgcn}. This is a characteristic of GCN, where gradients generally decrease as the hop increases. Furthermore, the features in T4 (orange) exhibit the smallest gradient values, suggesting that they are largely excluded during training and only receive minor updates through weight regularization. Figure \ref{gradient} describes the gradients in the shifted space. Contrary to the original space, all gradients have similar magnitudes with small deviations ($\sigma$) regardless of their types. This indicates that the proposed shifting mechanism has a potential to rectify the problem incurred by the initial feature sparsity by updating most dimensions during training.

\begin{table}[h!]
\centering
\caption{(RQ5) Ablation study on four graphs showing the node classification accuracy (\%) of seven Shift-GCN variants}
\label{ablation2}
\begin{adjustbox}{width=.48\textwidth}
\begin{tabular}{l|c|c|c|c}
\hline
\textbf{Category}    & \textbf{Cora} & \textbf{Citeseer} & \textbf{Actor} & \textbf{Cornell} \\ \hline
(a) shift $1^{st}$ plane            & 77.4                       & 66.6 & 20.1 & 36.8                           \\ \hline
(b) shift $2^{nd}$ plane           & 80.1                       & 71.8 & 22.8 & 41.5                           \\ \hline
(c) set as $s=1$ (Eq. \ref{flip_init})        & 82.3                       & 72.1 & 27.6 & 47.4                           \\ \hline
(d) w/o iterative train.        & 82.0                       & 71.4 & 28.2 & 48.0                           \\ \hline
(e) w/o plane shift           & 81.7                       & 70.0 & 23.5 & 41.3                           \\ \hline
(f) w/o early stop           & 82.5                       & 72.5 & 28.8 & 49.2                           \\ \hline
\textbf{Original Shift-GCN}         & \textbf{82.9}                       & \textbf{72.6} & \textbf{29.1} & \textbf{49.4}                          \\ \hline
\end{tabular}
\end{adjustbox}
\end{table}

\subsection{(RQ5) Ablation Study}
In Table \ref{ablation2}, we conduct ablation studies on Shift-GCN by systematically varying each component of the module: (a) shifting only the first hyperplane, (b) applying shift to the second layer instead of the first hyperplane, (c) setting the shifted value in Eq. \ref{flip_init} as $s=1$, (d) training the model only in a shifted space, (e) shifting only the initial features, (f) excluding early stopping from the original model, and using the baseline Shift-GCN configuration. The results indicate that removing zero elements is crucial, as (e) shifting the features alone outperforms (a) hyperplane shift. The simultaneous shifting of the first plane is also important, as (b) shifting only the second layer results in lower accuracy than (a). Notably, (d) training entirely in a shifted space shows competitive accuracy, suggesting that non-zero elements significantly contribute to the performance. Furthermore, (f) early stopping after 50 iterations maintains good accuracy and results in faster convergence, as supported by Figure \ref{convergence}. Interestingly, we noticed that perturbing the initial features leads to significant performance improvement, along with the original Shift-GCN. This result can be attributed to reduced prediction variance achieved through shifting.

\section{CONCLUSION}
Most previous GNNs focus on refining aggregation strategies, often overlooking the nature of initial feature types. This study investigates how zero elements in input vectors influence the optimization of the first layer. We propose a co-training method that learns gradient flows across the original and shifted spaces, dynamically adapting parameters throughout the process. We also offer a theoretical explanation showing that shifting reduces prediction variance leading to the stable convergence. While the improvements for datasets without zero elements are modest, our method yields significant enhancements for other types of graphs, highlighting the scalability and effectiveness of our approach.

\section{Acknowledgments} 
%This work was supported in part by KENTECH Research Grant (202200019A) and in part by Institute of Information \& Communications Technology Planning \& Evaluation (IITP) grant funded by the Korean government (MSIT) (No. 2021-0-02068, IITP-2024-00156287).
This work was supported by the KENTECH Research
Grant (202200019A), by the Institute of Information \& Communications Technology Planning \& Evaluation(IITP) grant funded by the Korea government(MSIT) (IITP-2024-RS-2022-00156287), and
by Institute of Information \& Communications Technology Planning \& Evaluation(IITP) grant funded by the Korea government(MSIT) (No.RS-2021-II212068).

\bibliographystyle{ACM-Reference-Format}
\balance
\bibliography{references.bib}

\end{document}